\documentclass[10pt,twocolumn,letterpaper]{article}

\title{Evaluation of Human Visual Privacy Protection: A Three-Dimensional Framework and Benchmark Dataset}

\author{Sara Abdulaziz\\
Eindhoven University of Technology\\
Eindhoven, The Netherlands\\
{\tt\small s.e.a.m.abdulaziz@tue.nl}
\and
Giacomo D'Amicantonio\\
Eindhoven University of Technology\\
Eindhoven, The Netherlands\\
{\tt\small g.d.amicantonio@tue.nl}
\and 
Egor Bondarev\\
Eindhoven University of Technology\\
Eindhoven, The Netherlands\\
{\tt\small e.bondarev@tue.nl}
}
\usepackage[pagenumbers]{iccv} 
\usepackage{graphicx}%
\usepackage{multirow}%
\usepackage{amsmath,amssymb}%
\usepackage{amsthm}%
\usepackage{mathrsfs}%
\usepackage{xcolor}%
\usepackage{textcomp}%
\usepackage{manyfoot}%
\usepackage{booktabs}%
\usepackage{algorithmicx}%
\usepackage{algpseudocode}%
\usepackage{listings}%
\usepackage{makecell}
\usepackage{longtable}
\usepackage{float}
\usepackage[export]{adjustbox}
\usepackage{array}
\usepackage{paralist}
\usepackage{setspace}
\usepackage{soul}
\usepackage[utf8]{inputenc}
\usepackage{caption}
\usepackage{subcaption}
\usepackage[pagebackref,breaklinks,colorlinks,allcolors=iccvblue]{hyperref}
\usepackage{enumitem}    

\definecolor{iccvblue}{rgb}{0.21,0.49,0.74}

\newcommand{\best}[1]{\textcolor{red}{\textbf{#1}}}
\newcommand{\second}[1]{\textcolor{blue}{\textbf{#1}}}

\def\paperID{*****} 
\def\confName{ICCV}
\def\confYear{2025}

\begin{document}
\maketitle

\begin{abstract}

Recent advances in AI-powered surveillance have intensified concerns over the collection and processing of sensitive personal data. In response, research has increasingly focused on privacy-by-design solutions, raising the need for objective techniques to evaluate privacy protection. This paper presents a comprehensive framework for evaluating visual privacy-protection methods across three dimensions: privacy, utility, and practicality. In addition, it introduces HR-VISPR, a publicly available human-centric dataset with biometric, soft-biometric, and non-biometric labels to train an interpretable privacy metric. We evaluate 11 privacy protection methods, ranging from conventional techniques to advanced deep-learning methods, through the proposed framework. The framework differentiates privacy levels in alignment with human visual perception, while highlighting trade-offs between privacy, utility, and practicality. This study, along with the HR-VISPR dataset, serves as an insightful tool and offers a structured evaluation framework applicable across diverse contexts.

\end{abstract}    
\section{Introduction}
\label{sec:intro}



Recent AI advancements have enabled intelligent monitoring systems capable of real-time event detection and behavioral analysis \cite{AD_crowd}. However, these systems raise significant privacy concerns due to the extensive collection and processing of private data. Consequently, data protection regulations, such as the GDPR \cite{GDPR}, have mandated privacy-by-design strategies, driving extensive research in visual privacy protection \cite{35_climent2021protection,AD10_fioresi2023ted,79_gao2023privacy,HAR16rp_hinojosa2021learning}.
As visual privacy-protection methods inherently cause information loss through the obfuscation of sensitive regions, they often degrade the performance of downstream computer vision tasks, i.e., utility tasks. Even though utility models should ideally be independent of sensitive human attributes, such as face, skin-color, or clothing, performance drops presist even when models are re-trained on anonymized data \cite{AD10_fioresi2023ted, HAR30_dave2022spact, AD4_cucchiara2024video, AD1_mishra2023privacy}. 





Prior work on the privacy-utility trade-off analysis for human-related vision tasks often targets narrow protection methods \cite{AD4_cucchiara2024video, AD5_yan2020image}. 
For instance, in \cite{AD4_cucchiara2024video}, the authors assess the utility performance across the protection degree of a single method (e.g., blurring). In \cite{AD5_yan2020image}, the performance is evaluated across the degree of sensitive object removal (e.g., bounding-box vs. segment). Other works, which include a broader assessment of protection methods, focus either on image-level, private vs. public \cite{zhao2022privacyalert}, or on face-level de-identification \cite{2017-Erdelyi}. Despite offering insights into the utility performance under narrow protection methods, they remain insufficient to evaluate complex protection approaches.


Recently, trainable privacy metrics have gained attention with the introduction of multi-label visual privacy datasets, such as VISPR \cite{VISPR}. The primary approach involves training a privacy-attribute classifier on the unprotected private data, then computing the class-based mean average precision on the protected test sets to rank the privacy-protection level \cite{VISPR,AD10_fioresi2023ted,HAR30_dave2022spact,PA-HMDB51}. Despite its potential, the reliability of this approach is compromised by several shortcomings in prior implementations \cite{HAR3_wu2020privacy,AD10_fioresi2023ted,HAR30_dave2022spact,PA-HMDB51}. These include the use of inaccurate privacy-labeled data and evaluation on a narrow subset of privacy attributes and protection methods.




In this paper, we propose an objective three-dimensional privacy evaluation framework assessing visual privacy-protection methods across privacy, utility, and practicality. It employs a human-centric privacy metric trained on HR-VISPR, a newly introduced dataset with comprehensive human-related labels, including biometric, soft-biometric, and non-biometric attributes. HR-VISPR serves as a benchmark for future research. Using this framework, we evaluate 11 visual privacy-protection methods, ranging from traditional to advanced deep-learning based methods, offering a more robust and explainable assessment of the privacy-utility-practicality trade-off.


\section{Related Work}\label{sec:related-work}

Privacy-protection evaluation spans two domains. The first, focused on guiding online data-sharing decisions \cite{VISPR,PrivAttNet,PM-1,PM-2}, lies outside our scope. The second evaluates privacy-by-design solutions across utility domains \cite{AD10_fioresi2023ted,2017-Erdelyi,PM-3,PM-7}, with schemes tailored to specific tasks. These evaluation methods primarily assess two aspects: privacy protection level and the utility performance of the anonymized data.

\noindent \textbf{Face-Level Privacy Evaluation.} In the domain of face recognition, privacy protection involves altering facial details either by face synthesis \cite{9_gafni2019live,4_hukkelaas2019deepprivacy}, or face perturbation \cite{100_chandrasekaran2020faceoff,62_low2022adverfacial}. In this context, the objective of privacy evaluation is to ensure anonymity against AI-based recognition models. Several metrics are employed to measure the face protection level, including face de-identification rate \cite{2017-Erdelyi,11_wu2019privacy,70_maximov2022decoupling,71_hukkelaas2023realistic}, face-shape retention rate \cite{korshunov2013framework}, face verification \cite{38_lee2021development}, k-anonymity score \cite{23_nousi2020deep,40_croft2021obfuscation}, and user survey metrics \cite{9_gafni2019live,17_sattar2020body,HAR40_wang2023modeling,100_chandrasekaran2020faceoff}. 


\noindent \textbf{Human-Level Privacy Evaluation.} When the entire human body is targeted for privacy protection, it causes a greater loss of visual information, leading to a higher performance drop in utility tasks that rely on human features, such as action recognition and anomaly detection. One approach to evaluate the protection level in this case is by assessing human recognition scores in user surveys \cite{HAR40_wang2023modeling}. However, to avoid their time-consuming nature and the potential bias, recent methods employ deep learning models in place of human evaluators, measuring privacy attribute prediction accuracy \cite{AD10_fioresi2023ted, HAR30_dave2022spact,43_sepehri2021privacy, HAR3_wu2020privacy}. A common approach trains a privacy-attribute classifier on non-anonymized visual data to detect human attributes like face, hair color, race, and gender \cite{HAR3_wu2020privacy}. The classifier is then applied to anonymized data, and the class-based mean average precision (cMAP) is computed to quantify the privacy level of each anonymization method. To analyze the privacy-utility trade-off, the anonymized data is applied to a utility task, and the privacy-utility scores are analyzed to help in selecting the best privacy-protection method \cite{AD10_fioresi2023ted,HAR30_dave2022spact}.
\section{Privacy Evaluation Framework} \label{sec:framework}
This section presents the human-centric visual privacy dataset and the privacy evaluation framework. 
 
\subsection{HR-VISPR Dataset} \label{sec:HR-VISPR}
\begin{figure*}[htb]
    \centering
    \includegraphics[width=\linewidth]{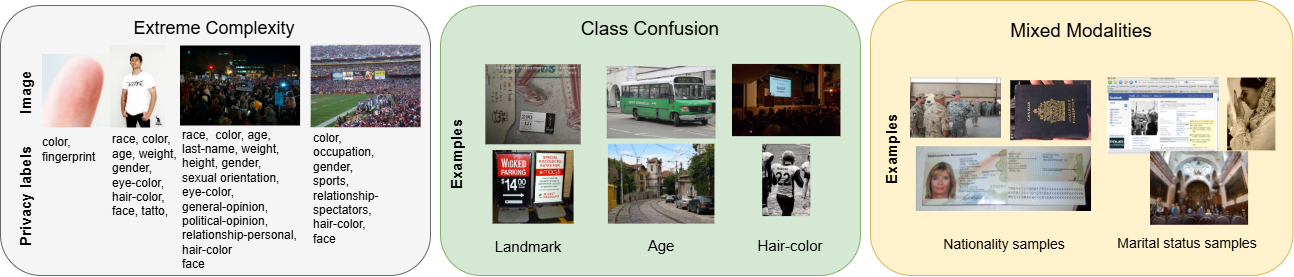}
    \caption{Examples of VISPR dataset limitations. }
    \label{fig:VISPR-problems}
\end{figure*}

Existing visual privacy datasets vary widely in their annotation type due to the differing research assumptions. The common and popular privacy datasets include PicAlert \cite{zerr2012picalert}, YourAlert \cite{youralert_spyromitros2016personalized}, VizWiz-Priv \cite{Vizwiz-priv_gurari2019vizwiz}, PrivacyAlert \cite{zhao2022privacyalert}, DIPA \cite{DIPA_xu2023dipa}, BIV-Priv \cite{BIV-Priv_sharma2023disability}, and DIPA2 \cite{DIPA2_xu2024dipa2}. These datasets are mostly annotated with a binary image label, e.g. private or public, according to the presence of personally identifiable information. In contrast, the Visual Privacy dataset (VISPR) \cite{VISPR}, proposed initially for guiding online image sharing decisions, incorporates a wide range of human-related privacy attributes. It consists of 22k personal and public images, annotated using 68 binary privacy attributes, including face, gender, age, nudity, hobbies, race, skin color, ethnic clothing, relationships, ID documents, address, and so on. Due to the wide coverage of private attributes, the dataset has been serving as a useful benchmark for privacy-protection evaluation \cite{AD10_fioresi2023ted, HAR30_dave2022spact,PA-HMDB51}. However, despite its diversity, the dataset has several major limitations that significantly impact its fairness for the privacy evaluation purposes. These limitations, illustrated in Fig.\ref{fig:VISPR-problems}, include the following.


\noindent \textbf{{Extreme Complexity.}} the dataset combines a wide range of complex scenes, including distant crowd scenes, wide-view images with humans represented by a tiny fraction of pixels, selfie images, body-part images, and poorly illuminated images. However, this complexity is often overlooked during the annotation process, making it challenging for a model to learn feature-to-label associations effectively, particularly given the large number of labels.

\noindent\textbf{{Class Confusion.}} the dataset includes a significant number of noisy samples, e.g., images assigned irrelevant privacy attributes. Such examples disturb the learning process, as the model will likely learn incorrect associations between labels and image content. 

\noindent\textbf{{Mixed Modalities.}} the dataset combines multiple modalities, including text-only, text with visual data, screenshots of online platforms, etc. The co-existence of both textual and visual data to represent a single privacy attribute can significantly impact the training. For example, learning the age attribute from both visual traits and text in IDs is extremely difficult and may lead to models that behave undesirably, misleading downstream tasks.

We propose a human-related visual privacy dataset (HR-VISPR), derived from VISPR, resolving the previously explained limitations and introducing complementary labels. In HR-VISPR, we cover human biometric, soft-biometric, and non-biometric attributes, enabling effective training of privacy-protection evaluation models. The dataset contains 10.11k samples with 18 privacy labels, derived via the following procedure from VISPR. More details on the attribute distribution are provided in \textcolor{blue}{Supp. S2.}

\noindent
\hangindent=3.6em 
\hangafter=1     
\textbf{Step 1:} We unified the modality to visual content only, excluding textual, document-like, samples.
\par 

\noindent
\hangindent=3.6em 
\hangafter=1
\textbf{Step 2:} We excluded confusing samples, where the image content is irrelevant to the attributes. Then, merged partial and complete attributes, such as partial and complete face and nudity.
\par

\noindent
\hangindent=3.6em
\hangafter=1
\textbf{Step 3:} Human-related biometric, non-biometric, and soft-biometric attributes were re-used. These include: `age', `face', `skin color', `hair color', `gender', `nudity', `height', `weight', `ethnic clothing', `religion', `race', `disability', and `sports'.
\par

\noindent
\hangindent=3.6em
\hangafter=1
\textbf{Step 4:} Soft-biometric attributes were extended to include clothing styles, among which are the two existing attributes: `sports' and `ethnic-clothing'. The following five new clothing labels were added: `casual clothing', `formal clothing', `uniforms', `troupe attire', and `medical scrubs'. Examples are shown in \textcolor{blue}{Fig. S1.}
\par

\noindent
\hangindent=3.6em
\hangafter=1
\textbf{Step 5:} Context correction for all re-used labels, where original assumptions by the authors differ from our research assumption. This includes removing samples with non-human objects or backgrounds as the main privacy indicators. For example, images of religious buildings with religion label, or images of sport equipment only with sports label, and so on.
\par

\subsection{Three-dimensional Evaluation Framework} \label{sec:privacy-utility-practicality}

\begin{figure*}
    \centering
    \includegraphics[width=\linewidth]{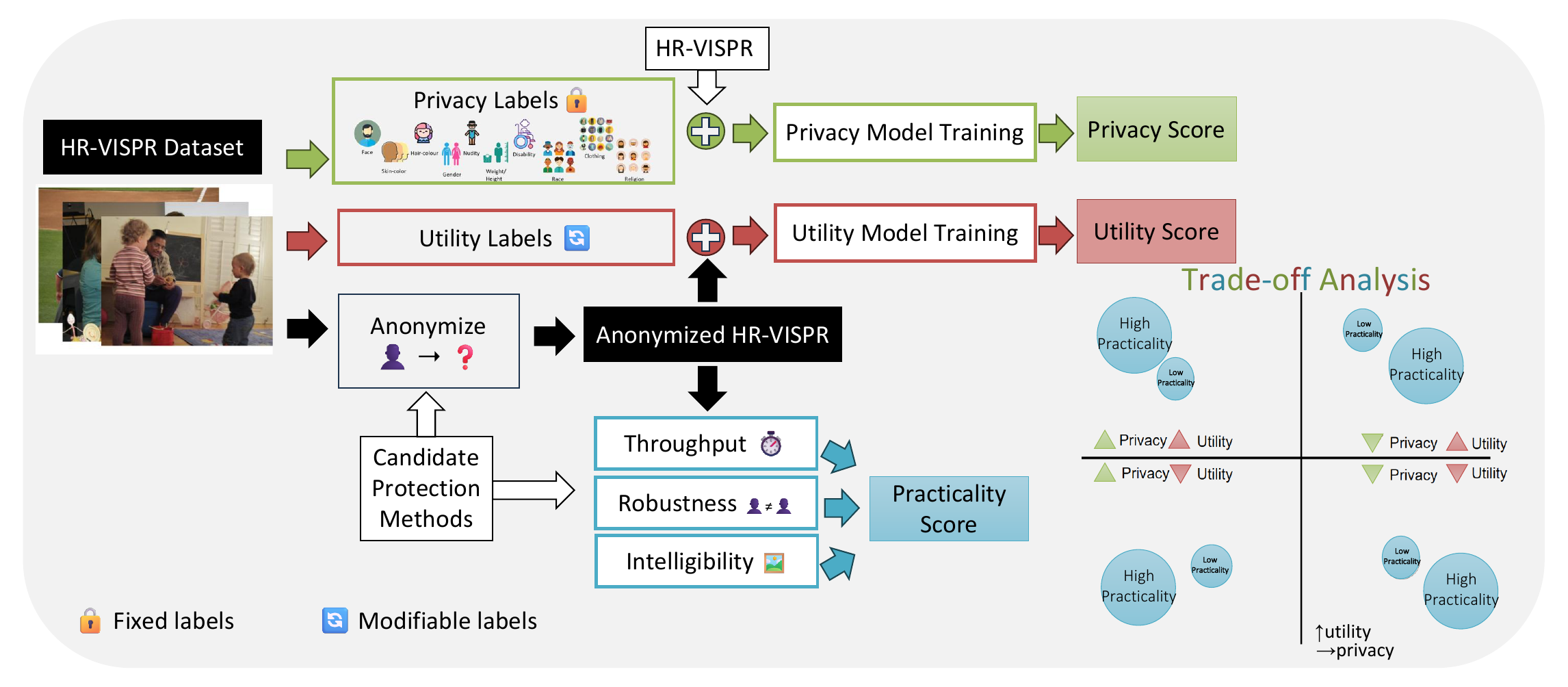}
    \caption{Overview of the three-dimensional approach for privacy-protection evaluation, based on HR-VISPR. Privacy, utility, and practicality are represented by the horizontal axis, vertical axis, and circle diameter, respectively. }
    \label{fig:3d-framework}
\end{figure*}

Our framework evaluates three key dimensions: privacy, utility, and practicality, which are the most critical metrics in evaluating privacy protection. Unlike prior evaluation schemes that merge privacy and utility scores \cite{2017-Erdelyi}, the framework treats these dimensions independently, providing a clearer reasoning when selecting a privacy-protection method. The evaluation follows a structured, adaptable procedure. First, candidate privacy-protection methods to be evaluated across the three dimensions are applied to the privacy-utility dataset, HR-VISPR. Second, the privacy classifier and utility model are trained on the dataset independently, ensuring that the utility model remains unbiased to the privacy-attribute features of the classifier. Finally, metrics are computed and the optimal protection method is selected. Fig. \ref{fig:3d-framework}
 shows the framework overview.

\subsubsection{Measuring Privacy} \label{sec:privacy-metric}

To assess privacy protection, we measure the performance degradation of a privacy-attribute classifier when applied to anonymized data, following prior works \cite{VISPR,AD10_fioresi2023ted,HAR30_dave2022spact,PA-HMDB51}. The classifier predicts a binary label for each privacy attribute based on its presence in the image. When the classification accuracy drops, it indicates a stronger anonymization, as effective privacy protection suppresses or distorts the sensitive features. The performance is quantified by the class-based mean average precision (cMAP), computed as \( \frac{1}{N} \sum_{c=1}^{N} \text{mAP}_c \), where \( N \) is the total number of privacy attributes and \( \text{mAP}_c \) denotes the mean average precision for attribute \( c \). The cMAP metric captures the classifier’s confidence in recognizing the different sensitive attributes. Low cMAP scores reflect a strong overall attribute protection. 

\subsubsection{Measuring Utility} \label{sec:utility_measure}
To assess the utility for each anonymization method, we employ object detection, a fundamental task in video surveillance, applied as follows. First, we construct the HR-VISPR utility labels, selecting the 10 most relevant objects aligning with the real-world surveillance context. These objects are chosen from the 80 object labels in the popular COCO dataset \cite{coco_lin2014microsoft}, namely: `person', `bicycle', `car', `motorcycle', `airplane', `bus', `train', `truck', `boat', and `traffic light'. We derive the ground truth labels from a pre-trained model \cite{yolov8}. More details about the object distribution are in \textcolor{blue}{Supp. S2.} Second, we train an object detector on the original and anonymized versions of HR-VISPR, obtained by candidate protection methods. Finally, the performance is assessed based on common performance metrics, such as precision, recall, AUC, and F1-score. While no anonymization method can completely prevent the utility degradation, effective methods are expected to be independent of human-sensitive attributes, minimizing their impact on utility.



\subsubsection{Measuring Practicality} \label{sec:practicality}
Practicality is a critical factor in privacy-preserving video monitoring, as it determines how effectively does an anonymization method operate in real-world settings. A practical anonymization method must balance real-time processing, effective suppression of sensitive content, and preservation of non-sensitive content quality. To measure practicality, we propose to fuse three factors: throughput, robustness, and intelligibility, computed as follows.

\noindent\textbf{Throughput.} Throughput quantifies the real-time feasibility of an anonymization algorithm, influenced by its computational complexity, image resolution, and hardware efficiency. We measure this factor in terms of processed frames per second (fps), by the following equation
\begin{equation} \label{eq:fps}
    \mathrm{FPS}=\frac{N}{\sum_{i=1}^{N-1}\left(t\left( \bar{v}^{i}\right)-t\left( \bar{v}^{i-1}\right)\right)},
\end{equation}

where \textit{N} is the number of processed frames, and $t()$ function is the processing time of a given image \textit{$\bar{v}$}. 

\noindent\textbf{Robustness.} We define the robustness of a privacy-protection algorithm as the ability to reliably anonymize human instances within a scene. This entails minimizing the false negative rate and maximizing the dissimilarity between original and anonymized objects, thereby eliminating the sensitive-object detectability after protection. In this framework, we propose to measure robustness by detecting and matching human instances with an object detector. The key idea is to evaluate whether the anonymization method fails to detect and anonymize humans or inadvertently preserves identifiable human details. Inspired by previous works \cite{2017-Erdelyi}, we apply a pre-trained object detector on HR-VISPR to detect humans on both the original and anonymized sets. The detections are then matched based on the Intersection over Union (IoU) score. We further introduce an SSIM-based matching, computing the similarity between the IoU-matched detections to ensure the dissimilarity of the anonymized and original human matches. The number of similar human matches forms the robustness score, where lower values indicate effective protection.

\noindent\textbf{Intelligibility.} Intelligibility of an anonymization algorithm, as defined by Badii et al. \cite{badii2014overview}, is the ability to anonymize sensitive attributes only, while retaining all other information in a video in order to maintain the usefulness of a monitoring system. Prior measures rely on visual quality metrics from user studies or ML models. \cite{badii2014overview}. We adopt CLIP Maximum Mean Discrepancy (CMMD) \cite{CMMD} as our intelligibility score, a recent metric that compares dataset distributions to evaluate generative quality. CMMD is well-suited for this task due to its robustness in capturing varying distortion levels, as well as its reliability across different sample sizes. We compute CMMD between the original and each of the 11 anonymized datasets.

\noindent\textbf{Computing Practicality.} The three metrics are combined into a single score by a weighted sum of normalized values. Before aggregation, we invert the robustness and intelligibility scores to ensure that higher values indicate better performance. The practicality score is computed as (\( w_r \cdot \mathrm{R_{n}} + w_i \cdot \mathrm{I_{n}} + w_t \cdot \mathrm{T_{n}} \)), where \( \mathrm{R_n} \), \( \mathrm{I_n} \), and \( \mathrm{T_n} \) are the normalized robustness, intelligibility, and throughput, respectively. The weights sum to unity and can be adjusted to emphasize the relative importance of each component.




\section{Experiments}\label{sec:experiments}
\begin{figure*}[htb]
\centering
\subfloat[HB]{\includegraphics[width=0.09\textwidth, height=0.10\textwidth]{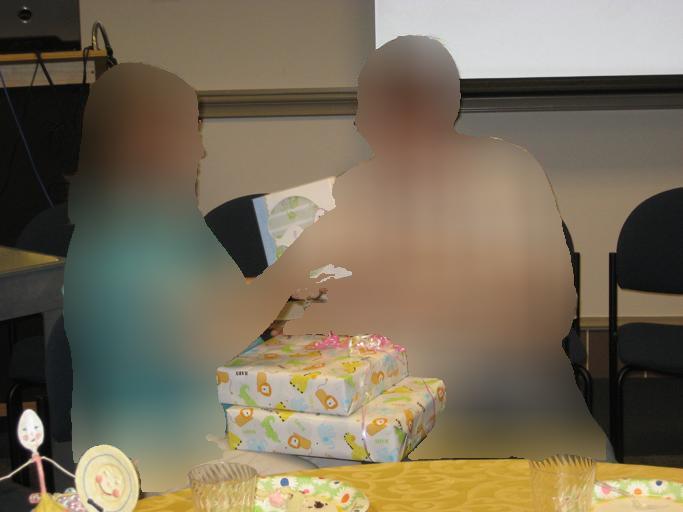}}
\subfloat[HP]{\includegraphics[width=0.09\textwidth, height=0.10\textwidth]{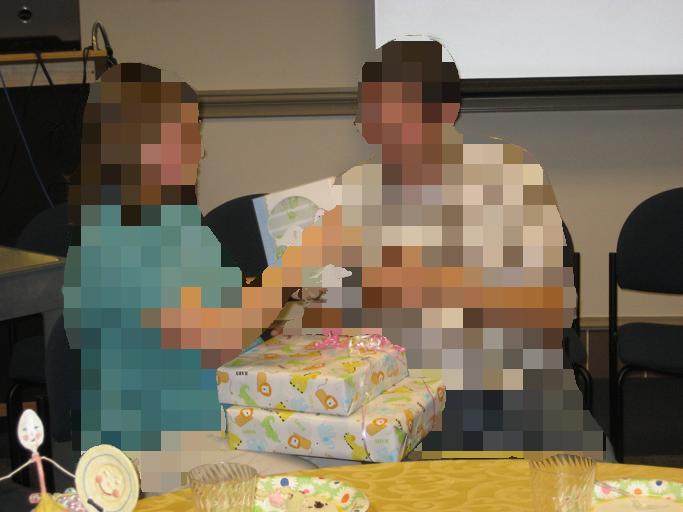}}
\subfloat[HE]{\includegraphics[width=0.09\textwidth, height=0.10\textwidth]{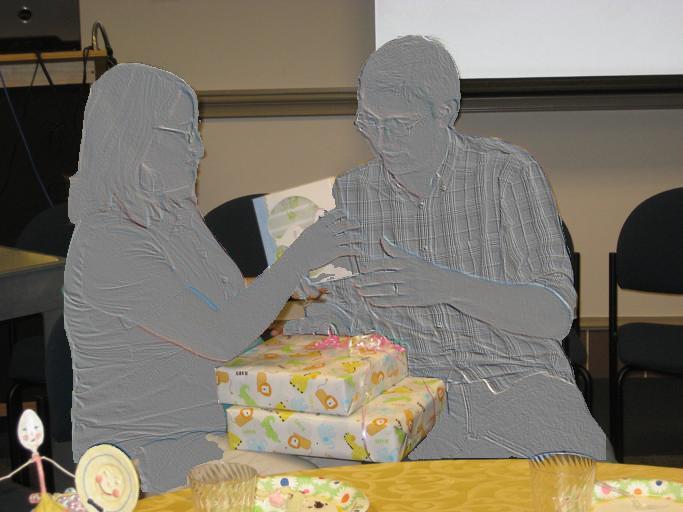}}
\subfloat[HM]{\includegraphics[width=0.09\textwidth, height=0.10\textwidth]{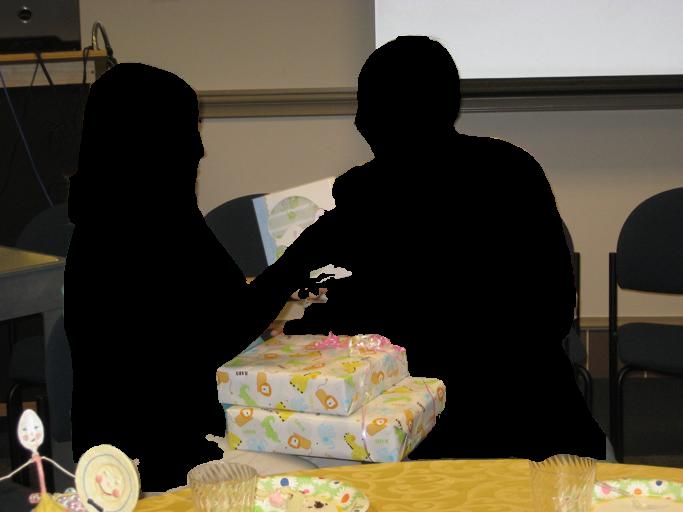}}
\subfloat[H2D]{\includegraphics[width=0.09\textwidth, height=0.10\textwidth]{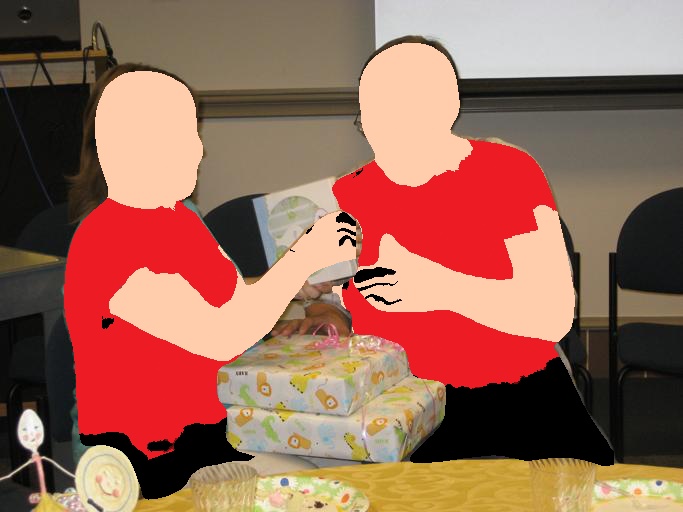}}
\subfloat[H3D]{\includegraphics[width=0.09\textwidth, height=0.10\textwidth]{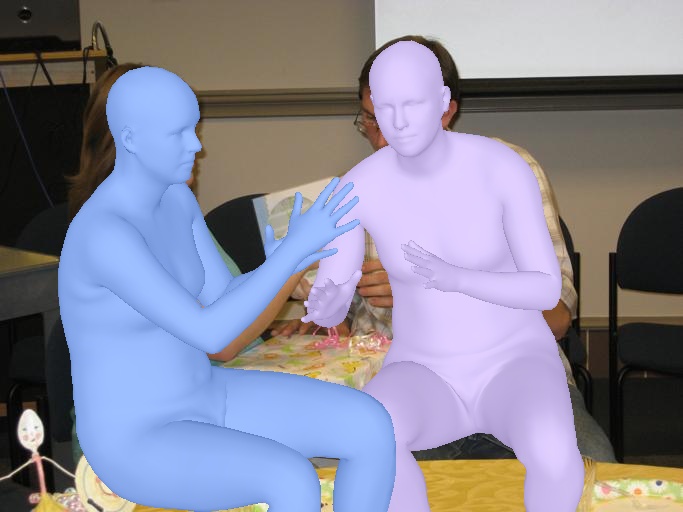}}
\subfloat[HEN]{\includegraphics[width=0.09\textwidth, height=0.10\textwidth]{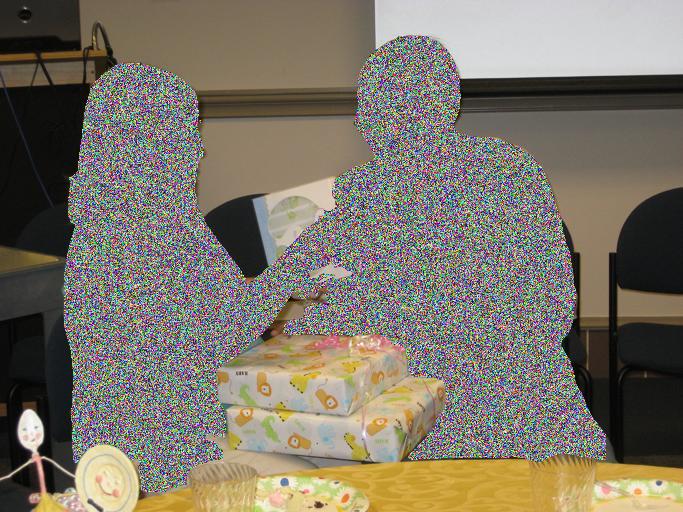}}
\subfloat[HS]{\includegraphics[width=0.09\textwidth, height=0.10\textwidth]{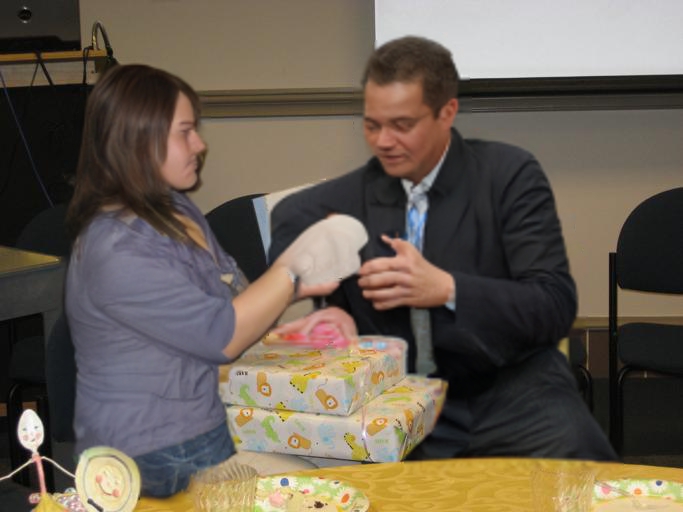}}
\subfloat[LR+SR]{\includegraphics[width=0.09\textwidth, height=0.10\textwidth]{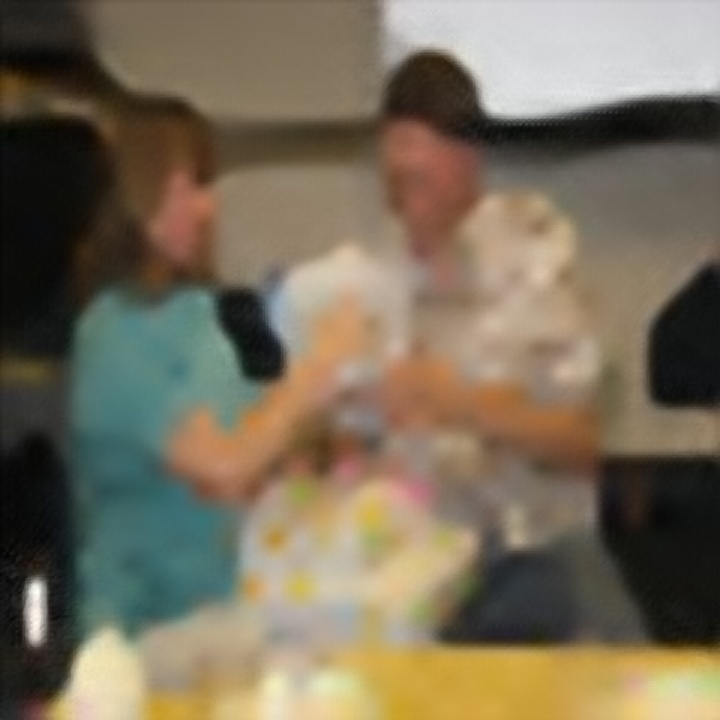}}
\subfloat[SPct]{\includegraphics[width=0.09\textwidth, height=0.10\textwidth]{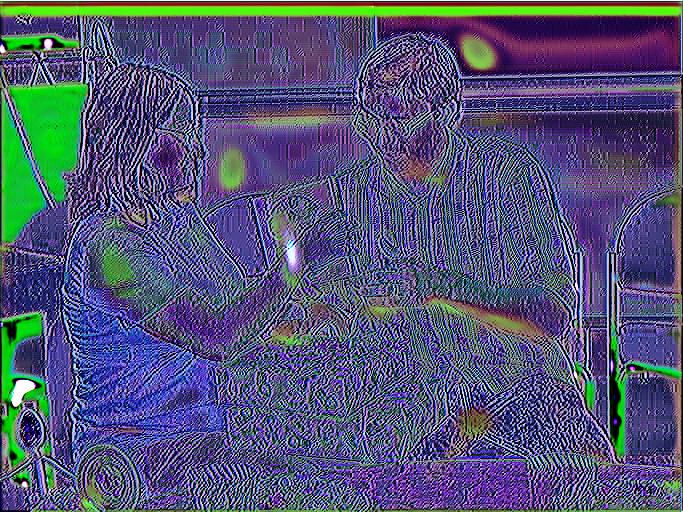}}
\subfloat[TSD]{\includegraphics[width=0.09\textwidth, height=0.10\textwidth]{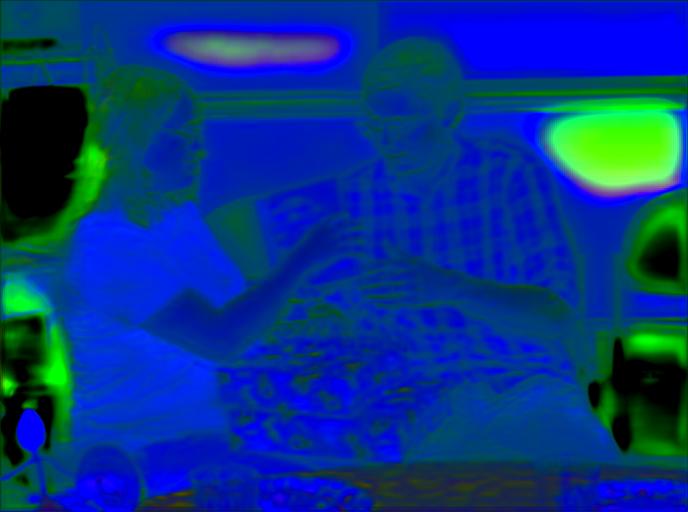}}
        
\caption{Visualizations of the anonymization methods applied in the evaluation framework, explained in \textcolor{blue}{Table S1.} }
\label{fig:visual-examples}
\end{figure*}

We evaluated a range of visual privacy-protection methods, including: Human Blurring (HB) \cite{82_tay2024privobfnet,HAR16_zhang2021multi}, Human Pixelation (HP) \cite{35_climent2021protection}, Human Embossing (HE) \cite{35_climent2021protection}, Human Masking (HM) \cite{AD5_yan2020image}, Human Encryption (HEN) \cite{14_shifa2020skin}, Human 2D-Avatar replacement (H2D) \cite{35_climent2021protection}, Human 3D-Avatar replacement (H3D) \cite{51_shen2023privacy,17_sattar2020body}, Human Synthesis (HS), Low Resolution + Super Resolution (LR+SR) \cite{HAR40_wang2023modeling}, SPAct (SPct) \cite{HAR30_dave2022spact}, and TeD-SPAD (TSD) \cite{AD10_fioresi2023ted}. The anonymized outputs of these methods are shown in Fig. \ref{fig:visual-examples}. Implementation details of the privacy-protection methods, as well as the privacy, utility, and practicality metrics, are provided in \textcolor{blue}{Supp. S3}.  




\subsection{HR-VISPR Dataset} The HR-VISPR dataset consists of 10.11k images, divided into training (7.11k), validation (1k), and testing (2k) sets. The training set includes an additional 788 safe images, which contain no human subjects (i.e., the multi-class labels are all set to 0), to improve the model generalization.

\subsection{Utility-Privacy-Practicality Evaluation}

\subsubsection{Evaluation of Privacy Metric Against Baseline} \label{exp1}
We evaluated the discriminative ability of the privacy classifier trained on HR-VISPR against a baseline aligned with prior works \cite{VISPR,AD10_fioresi2023ted, HAR30_dave2022spact,HAR3_wu2020privacy}. For the baseline, the same classifier is trained on original VISPR \cite{VISPR}, but only for a subset of seven labels: skin color, gender, partial face visibility, complete face visibility, partial nudity, personal relationship, and social relationship. This selection follows previous works \cite{HAR3_wu2020privacy, HAR30_dave2022spact, AD10_fioresi2023ted}, where these particular labels are considered the most prevalent across various datasets. The baseline training followed the same training configuration, in \textcolor{blue}{Supp. S3.1}, to ensure fair comparison. 

\begin{figure}[h!]
\centering
\includegraphics[width=\linewidth]{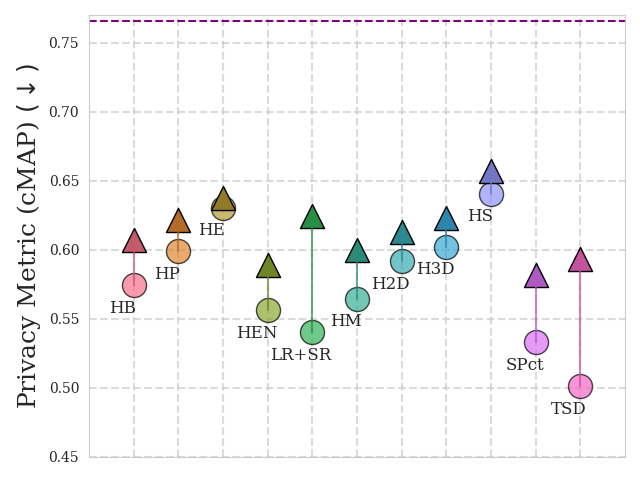}

\caption{Evaluation of the privacy-metric trained on HR-VISPR (circles) against the baseline trained on VISPR (triangles) \cite{HAR3_wu2020privacy,VISPR}. The purple dash line represents the former model performance on original images.  }
\label{fig:compare-to-baseline}
\end{figure}

Fig. \ref{fig:compare-to-baseline} compares the cMAP-based rankings obtained by the baseline and our HR-VISPR approach. Clearly, the privacy metric trained on HR-VISPR yields a broader privacy score range ($0.50-0.65$ cMAP), three times wider than the baseline ($0.60-0.66$ cMAP). This demonstrates the discriminative power of the privacy classifier when trained on an accurately constructed dataset, HR-VISPR. Additionally, the proposed rankings show better correlation with human perception than the baseline, when viewed in conjunction with Fig. \ref{fig:visual-examples}. For instance, the proposed approach assigns different privacy scores to LR+SR, H2D, and H3D, reflecting their varying levels of protection. In contrast, the baseline model assigns them nearly identical scores, indicating its limited discriminability. Similar insights can be drawn for methods (SPct and TSD) and (HM, LR+SR, and HEN).

\renewcommand{\arraystretch}{1}

\begin{table*}[hbtp!]
  \caption{Relative drop in mAP (\%) of the HR-VISPR classifier across attributes for each anonymization method. Higher drop indicates higher protection.
    {\color{red}\textbf{Bold red}} = highest drop;  
    {\color{blue}\textbf{Bold blue}} = second-highest drop.}
  \label{tab:privacy_drop_proposed}
  \centering
  \resizebox{\linewidth}{!}{%
      \begin{tabular}{lcccccccccccccccccc}
        \Xhline{1pt}
        \textbf{Method} & 
        \small{\textbf{Age}} &
        \small{\textbf{Face}} & 
        \small{\textbf{Skin}} & 
        \small{\textbf{Hair}} & 
        \small{\textbf{Gender}} & 
        \small{\textbf{Race}} & 
        \small{\textbf{Nudity}} & 
        \small{\textbf{Height}} & 
        \small{\textbf{Weight}} & 
        \small{\textbf{Disability}} & 
        \small{\textbf{Ethnic}} & 
        \small{\textbf{Formal}} & 
        \small{\textbf{Uniforms}} & 
        \small{\textbf{Medical}} & 
        \small{\textbf{Troupe}} & 
        \small{\textbf{Sports}} & 
        \small{\textbf{Casual}} & 
        \small{\textbf{Religion}} \\
        \hline
        HB   & \second{2.21}  & \best{5.74} & 1.54 & 3.04 & \second{1.10} & \best{3.98} & 52.9  & 16.22 & 13.83 & 6.21  & 30.28 & 49.67 & 37.16 & 20.45 & 5.35  & 29.24 & \second{24.84} & 40.56 \\
        HP   & 1.64         & 4.27          & \best{1.86} & 3.16 & 0.98 & 3.49  & 42.14 & 14.23 & 10.74 & 2.41  & 30.31 & 38.24 & 34.44 & 24.16 & 3.79  & 19.51 & 21.83 & 42.67 \\
        HE   & 0.86         & 2.59          & 1.60  & 2.77 & 0.79 & 2.81       & 33.32 & 10.71 & 6.61  & 7.49  & 26.54 & 30.64 & 27.25 & 13.81 & 9.01  & 20.57 & 16.36 & 29.88 \\
        HEN  & 1.66         & 4.60          & 1.42  & 4.08 & 0.80 & 2.62       & 55.84 & 10.65 & 11.36 & 2.20  & \best{38.27} & \second{51.32} & \second{38.52} & 45.82 & 7.26 & 32.65 & \best{26.10} & 42.43 \\
        LR+SR& 1.54         & 3.00          & 1.47  & 4.05 & 0.76 & 3.27       & 45.08 & \second{18.91} & 14.46 & \best{57.21} & \second{36.96} & 46.08 & 32.44 & 30.46 & 1.87  & 36.27 & 22.26 & \best{50.13} \\
        HM   & \best{2.31}& \second{5.52}   & \second{1.71}& \best{4.71} & 1.09 & \second{3.54}       & \best{59.38} & 15.00 & 12.56 & 9.79  & 26.89 & \best{54.91} & 37.90 & 24.86 & 8.43  & 27.11 & 23.69 & 42.77 \\
        H2D  & 1.04         & 2.60          & 0.96  & 2.54 & 0.67 & 2.16       & \second{56.89} & 10.81 & 8.61  & 9.79  & 24.45 & 49.93 & 35.98 & 24.60 & 7.05  & 22.81 & 21.81 & 30.87 \\
        H3D  & 0.90         & 2.24          & 1.01  & 2.47 & \best{1.16} & 2.23   & 54.12 & 11.72 & 7.80  & 9.81  & 26.18 & 44.63 & 36.75 & 15.83 & 7.16  & 16.96 & 20.94 & 32.54 \\
        HS   & 0.22         & 0.53          & 0.83  & 1.90 & 0.41 & 0.95       & 51.25 & 12.55 & 6.79  & -0.16 & 20.95 & 30.22 & 32.58 & 13.51 & 5.22  &  7.85 & 15.60 & 23.99 \\
        SPct & 1.15         & 2.07          & 1.10  & 3.11 & 0.97 & 2.13       & 45.33 & 15.11 & \second{14.75} & 39.96 & 36.57 & 49.36 & 36.61 & \second{47.73} & \second{10.19} & \second{41.56} & 24.18 & 47.08 \\
        TSD  & 1.27         & 2.81          & 1.62  & \second{4.24} & 0.81 & 2.37 & 44.03 & \best{19.85} & \best{18.07} & \second{53.82} & 36.22 & 47.12 & \best{38.90} & \best{72.93} & \best{11.25} & \best{50.35} & 23.43 & \second{47.26} \\
        \Xhline{1pt}
      \end{tabular}
      }
\end{table*}

\renewcommand{\arraystretch}{1}
\setlength\tabcolsep{2pt}
 \begin{table}[hbtp!]
    \centering
    \caption{Relative drop in mAP (\%) of the baseline classifier across VISPR-subset attributes for each anonymization method. Higher drop indicates higher protection. 
    {\color{red}\textbf{Bold red}} = highest drop;  
    {\color{blue}\textbf{Bold blue}} = second-highest drop.}
    \label{tab:privacy_drop_baseline}
    \resizebox{\columnwidth}{!}{%
    \begin{tabular}{lccccccc}
        \Xhline{1pt}
        \textbf{Method} & 
        \small{\textbf{Gender}}&
         \small{\textbf{FaceComp.}} & 
         \small{\textbf{FacePartial}} & 
         \small{\textbf{Nudity}} & 
         \small{\textbf{Skin}} & 
         \small{\textbf{PersonalRel.}} & 
         \small{\textbf{SocialRel.}} \\
        \midrule
      
        HB    & 0.75       & 7.86         & 1.56        & 36.33      & 0.30       & 13.63      & \second{19.00} \\
    HP    & 0.41       & 6.04         & 1.92        & 28.59      & 0.25       & 15.18      & 16.82          \\
    HE    & 0.70       & 4.08         & 1.61        & 25.30      & -0.01      &  9.79      & 16.51          \\
    HEN   & 0.77       & \second{10.47} & 1.27      & \best{43.79} & 0.24     & 15.00      & \best{20.39}   \\
    LR+SR & 0.66       & 5.85         & 5.61        & 24.09      & 0.68       & 13.48      & 16.98          \\
    HM    & 0.46       & 7.31         & 1.19        & 41.52      & 0.30       & 15.17      & 18.49          \\
    H2D   & 0.86       & 8.32         & 0.87        & 35.17      & 0.59       & 12.18      & 17.12          \\
    H3D   & 0.51       & 3.16         & 0.58        & \second{42.14} & 0.04   & 10.50      & 11.49          \\
    HS    & 0.15       & 2.06         & -0.02       & 28.28      & 0.04       &  8.57      &  5.37          \\
    SPct  & \best{1.09}& \best{11.04} & \second{9.70} & 40.90    & \second{0.92}& \best{16.25}& 17.02         \\
    TSD   & \second{0.98}& 9.08       & \best{10.13} & 37.88     & \best{0.93}& \second{16.10}& 13.94       \\
        \Xhline{1pt}
    \end{tabular} 
    }
\end{table}

            

To assess the interpretability of the HR-VISPR-trained classifier, we analyze the relative drop in mAP for each privacy attribute across anonymization methods, as shown in Tables \ref{tab:privacy_drop_proposed} and \ref{tab:privacy_drop_baseline}. This drop, calculated as the difference (in percentage) in mAP between original and anonymized images, reflects the effectiveness of each method in suppressing specific attributes. Table \ref{tab:privacy_drop_proposed} demonstrates that the HR-VISPR-based model yields interpretable results aligned with human intuition. For example, methods that eliminate body parts completely, such as HM, HB, and HEN, show the highest drop for the relevant attributes, such as face, skin color, hair color, and nudity, which is justifiable. However, these same methods show smaller drops for contextual attributes, such as uniforms, sportswear, medical scrubs, as the model can establish correlations between the labels and contextual scene information that is not modified by anonymization methods. For instance, sportswear label is positive when the background is a basketball arena, and medical scrubs are predicted when the image shows hospital beds or an operation theater. In contrast, methods that anonymize the entire image, such as SPct, TSD, and LR+SR, result in higher suppression (higher drop in mAP) of both clothing-related and contextual attributes, such as religion and disability. Likewise, H2D and H3D, which replace human figures with avatars, show moderate drops for body-related features, consistent with the partial retention of visual cues in avatar representations.

In contrast, the baseline model trained on VISPR exhibits less coherent behavior, as seen in Table \ref{tab:privacy_drop_baseline}. Remarkably, it shows consistently the highest protection (highest mAP drop) for TSD and SPct, despite revealing some facial details in their altered-color images. Methods like HM, which eliminate human regions entirely, exhibit low protection (low mAP drop) for sensitive facial, nudity, and skin-color attributes. Additionally, even though the skin-color attribute is well-protected in HM, HE, HEN, the baseline model shows significant drop in mAP only for TSD and SPct. This suggests that the model is biased toward spurious correlations in the training data, likely stemming from the noisy labeling in VISPR.

\subsubsection{Three-dimensional Trade-off Analysis} \label{exp2} 

\begin{figure}[htbp!]
\centering
\includegraphics[width=\linewidth]{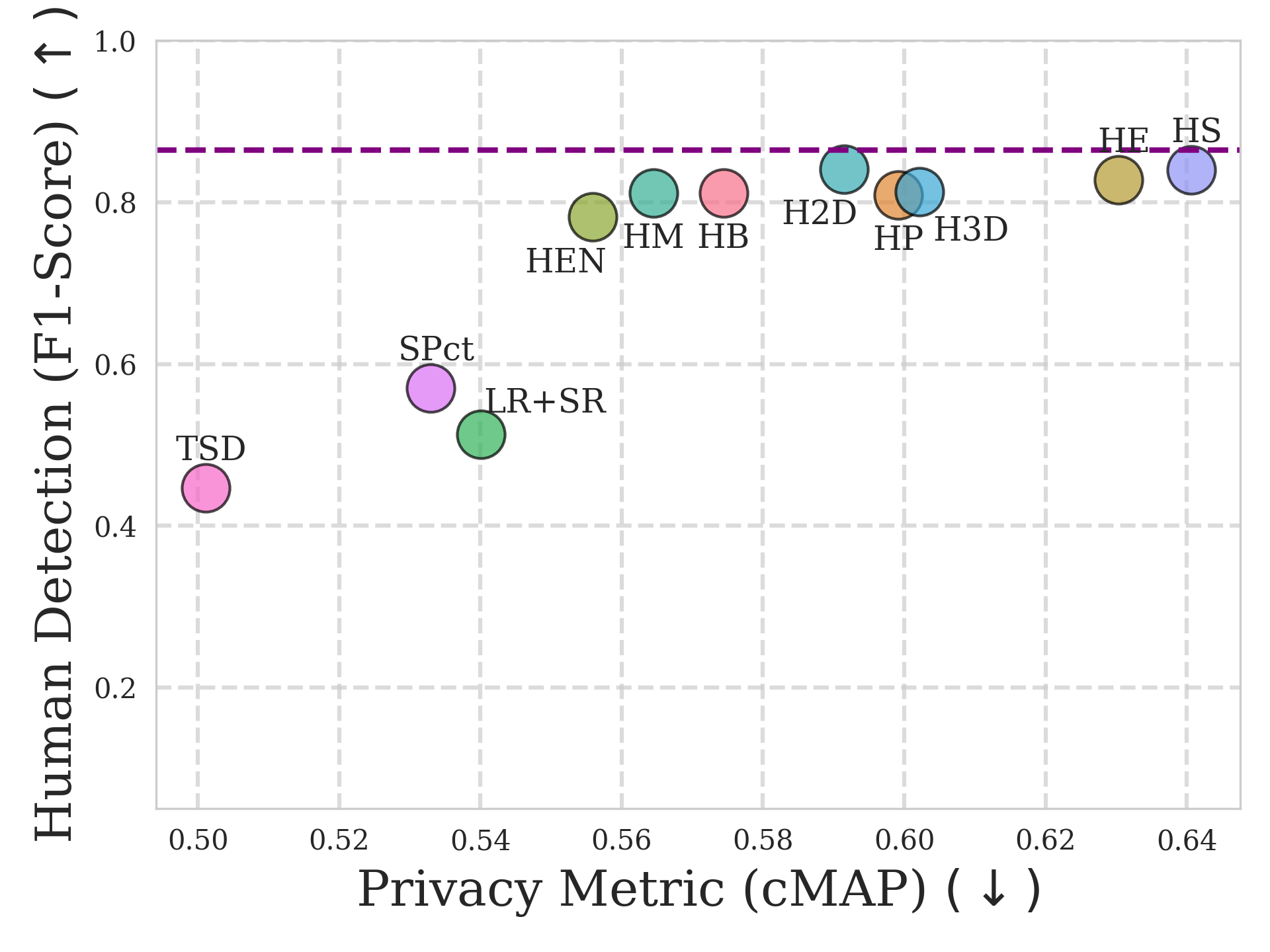}

\caption{The privacy-utility trade-off evaluation for the anonymization methods, with utility measured as the F1-Score of human detection on HR-VISPR dataset. The purple dash line represents the performance on the non-anonymized images.
}
\label{fig:2d-tradeoff}
\end{figure}

\noindent \textbf{Privacy Dimension.} Fig. \ref{fig:2d-tradeoff} shows the trade-off between privacy and utility. The privacy scores (horizontal axis) demonstrate the ranking of the anonymization methods, computed by the proposed privacy classifier. Analysing the scores in conjunction with the visualizations of these methods in Fig. \ref{fig:visual-examples}, we can see the correlation between the human perception and the score rankings. DL-based anonymization methods (SPct and TSD) and low resolution (LR+SR) achieve the highest protection levels ($0.5-0.55$ cMAP), as they obscure the entire image rather than selective human segments. 
Human encryption, masking, and blurring (HEN, HM, HB) provide intermediate protection ($0.55-0.59$ cMAP) due to exposing the scene context and relying on human body segmentation. The remaining methods (H2D, H3D, HP, and HE), occupy the lowest privacy range ($ 0.59-0.65 $ cMAP). While H2D and H3D may visually appear more preserving than HP and HE, they still reveal traces of body details due to poor human segmentation and imperfect body-to-avatar alignment. Additionally, they lack flexibility for the different clothing styles. On the other side, HP and HE reveal some human details which, when combined with the scene context, yield lower protection. Finally, HS shows the lowest privacy protection level ($ 0.64$ cMAP), closely matching the cMAP score of non-anonymized data, thereby validating the model's discriminative ability. Despite its low ranking, HS remains a strong privacy protection in reality due to synthesizing the entire human identities, see further discussion in Section \ref{sec:discussions}.


\noindent \textbf{Utility Dimension.} 
We evaluate the impact of anonymization on utility, starting with human detection and extending to other objects. The vertical axis in Fig. \ref{fig:2d-tradeoff} represents human detection performance on the anonymized HR-VISPR, measured by the average F1-score. Clearly, the overall performance drops under all anonymization methods, with HS, H2D, and HE maintaining nearly identical human detection to the original (non-anonymized) HR-VISPR. This is likely due to the replacement with rich features, such as synthetic human (HS), avatars (H2D), or the highlight and shadow effects (HE). LR+SR, TSD, and SPct yield the lowest performance, indicating limited applicability for object detection. Notably, deep-learning methods, such as TSD and SPct, designed for privacy-by-design human behavior analysis, retain only task-relevant motion cues, reducing their utility for general vision tasks. Other methods, such as HP, HB, HM, and H3D, show moderate degradation, validating their applicability for human detection under anonymization.

\begin{figure}
    \centering
    \includegraphics[width=\linewidth]{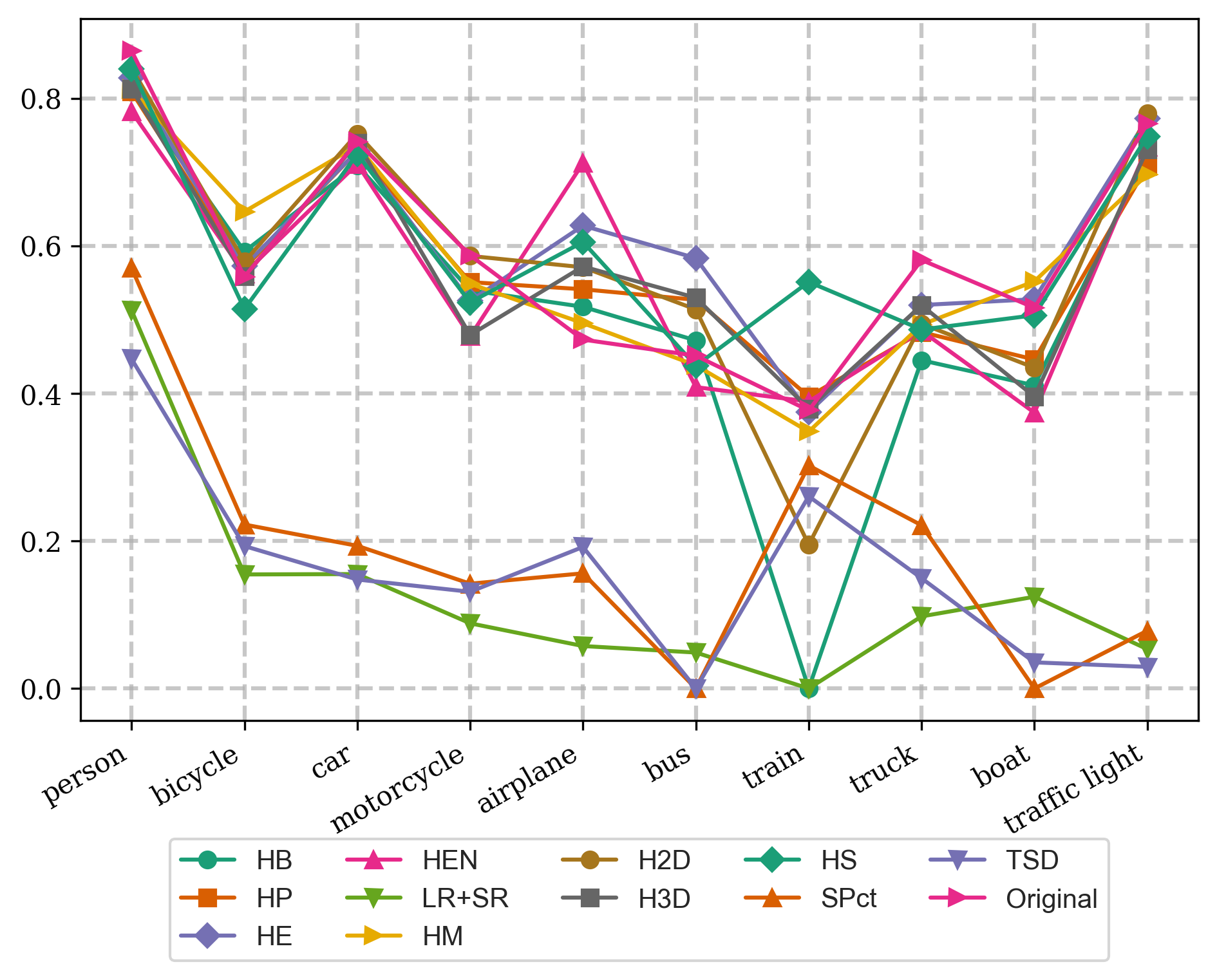}
    \caption{Utility F1-scores for the HR-VISPR objects.  }
    \label{fig:utility-f1}
\end{figure}

A similar observation is found for the detection performance of the other objects, as shown in Fig. \ref{fig:utility-f1}. Even though human anonymization introduces noise to the images, object detection remains feasible under most anonymization methods, with an F1-score close to the original, except for TSD, SPct, and LR+SR. The observed overall decline in the detection is attributed to the complexity introduced by the anonymized regions. For instance, specific objects, such as bicycles and motorcycles, mostly exhibit occlusions with anonymized regions (humans), which significantly impact performance due to the disturbed features. On the contrary, the performance on objects such as airplanes, buses, and traffic lights is shown to be higher than the original performance. This seems to arise from the imbalance between anonymized and non-anonymized image regions in the dataset. As most HR-VISPR images focus primarily on humans, this results in significant anonymized regions and information loss. In contrast, images without humans or with fewer human regions, as with the objects above, remain rich in detail, allowing the model to learn more effectively. This observation highlights how anonymization may contribute to dataset imbalance, suggesting that preparing utility datasets should address these potential pitfalls to ensure effective object detection performance. Further analysis of the utility is presented in \textcolor{blue}{Supp. S4.}

\begin{figure*}[htbp!]
    \centering
    \subfloat[Throughput emphasized]{\includegraphics[width=0.335\textwidth]{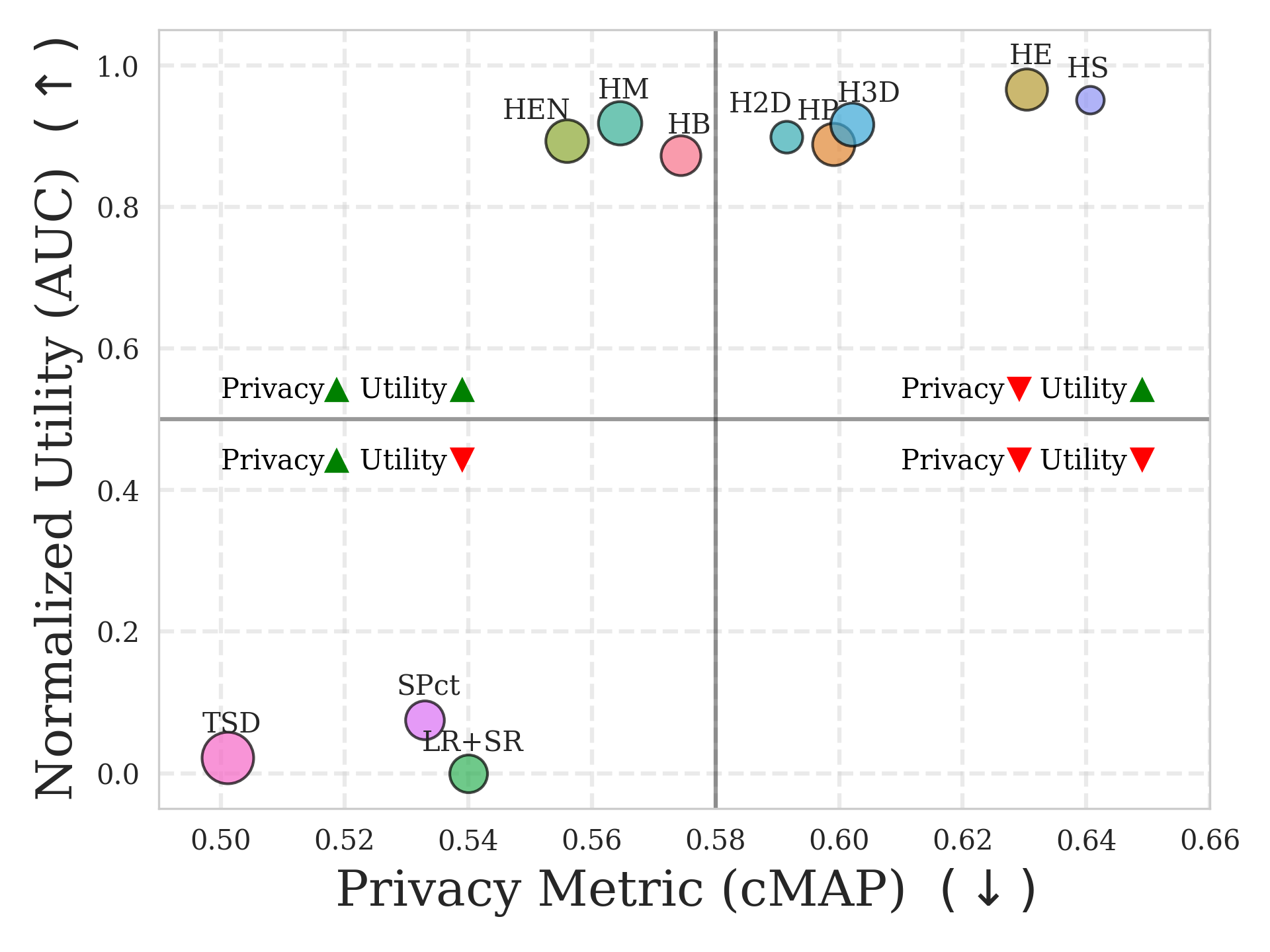}}
     \subfloat[Robustness emphasized]{\includegraphics[width=0.335\textwidth]{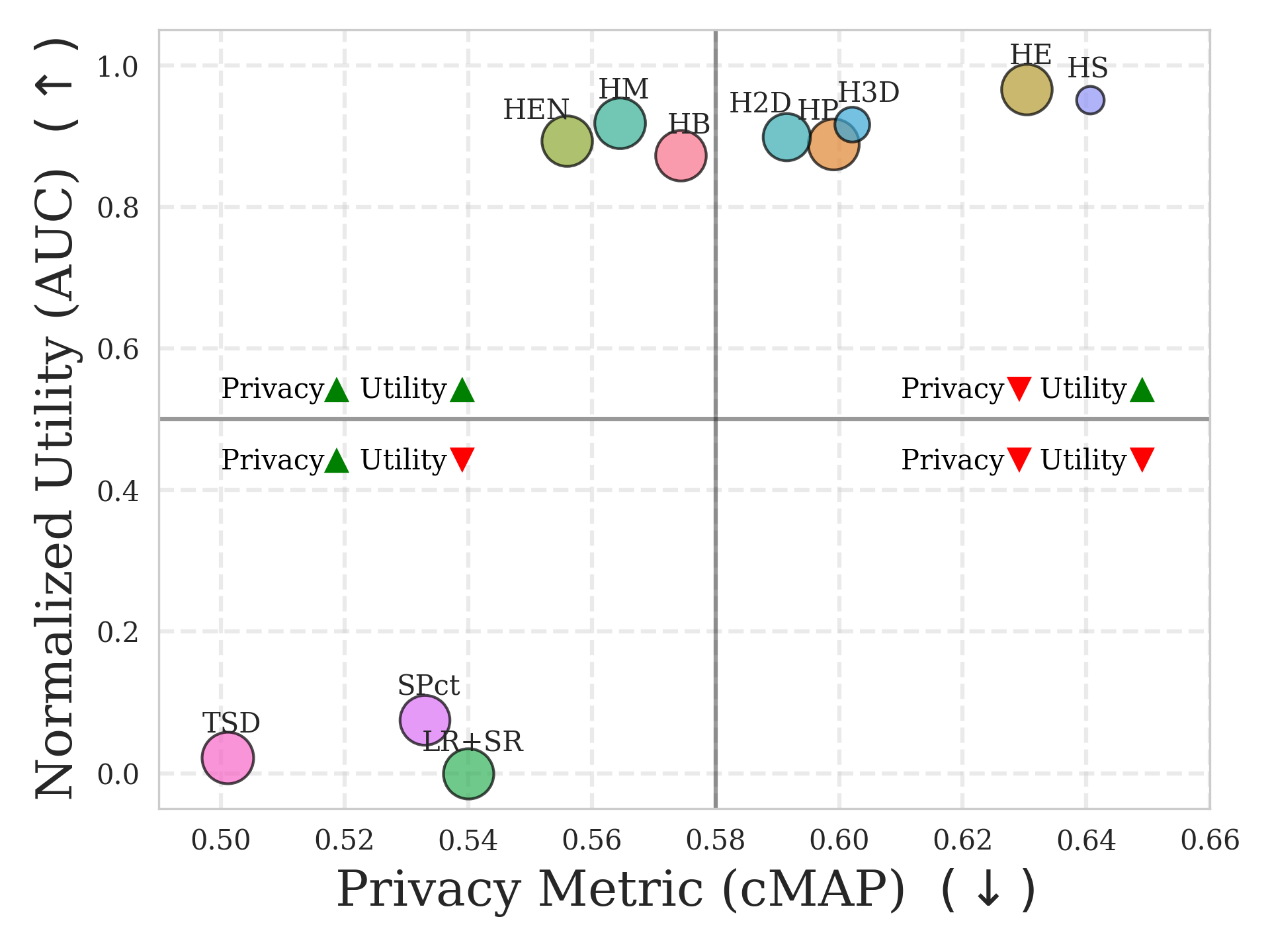}}
     \subfloat[Intelligibility emphasized] {\includegraphics[width=0.335\textwidth]{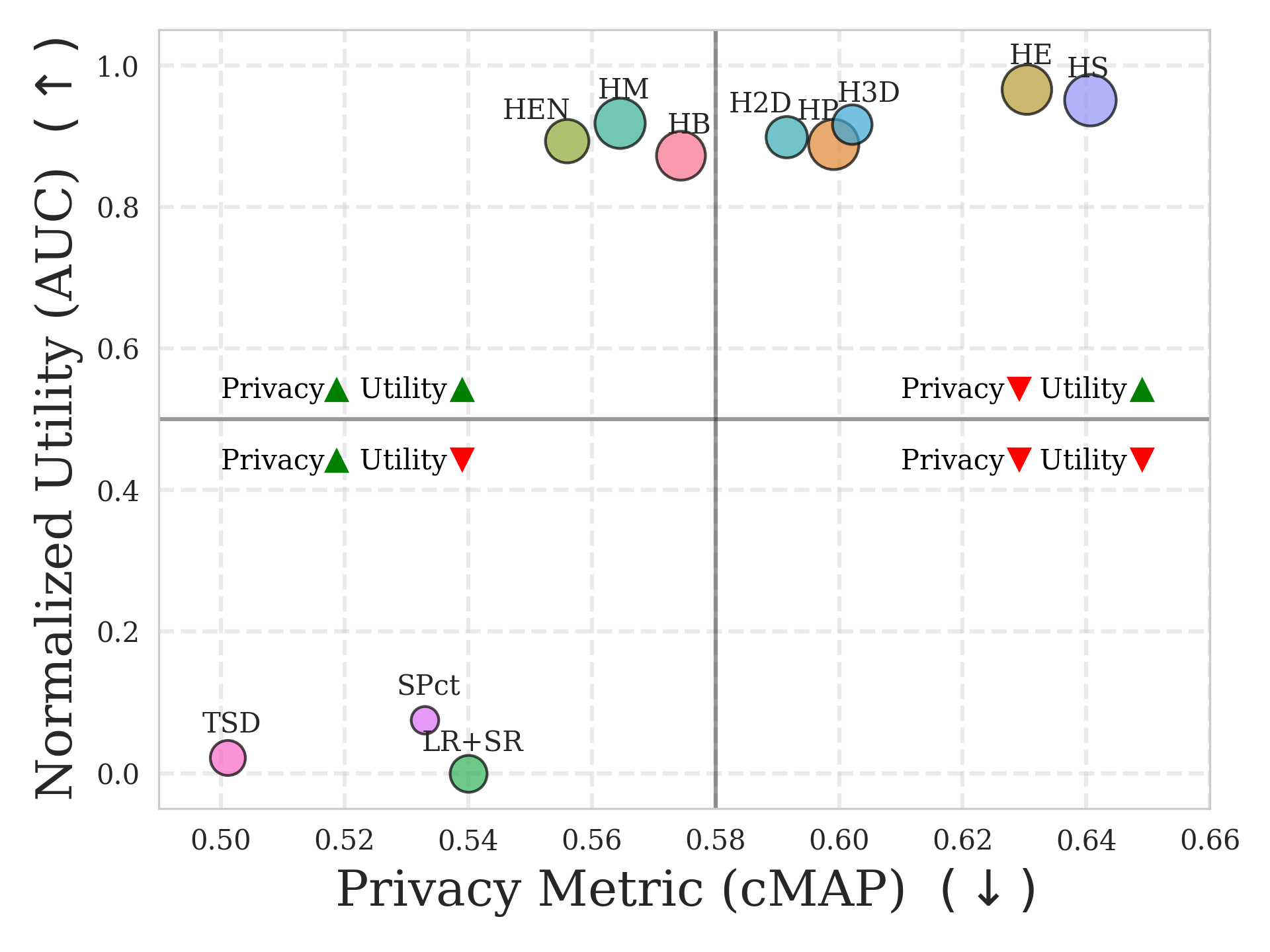}}
    
    \caption{The three-dimensional trade-off analysis of privacy, utility, and practicality. The third dimension (practicality) is represented by the circle diameter, where a larger diameter shows higher practicality.  }
    \label{fig:3d-graph}
\end{figure*}

\noindent \textbf{Practicality Dimension.} As the proposed framework aims to offer an explanatory tool that enables straightforward analysis of the protection methods, we analyze the practicality in three-dimensional trade-off plots, shown in Fig. \ref{fig:3d-graph}. The practicality is represented by the circle diameter, where larger circles indicate higher practicality. In each plot, one factor is emphasized by assigning it a weight of 0.8, while the others are set to 0.1. The plots in the figure categorize the anonymization methods into four groups based on their privacy-utility trade-off. The top-ranked group (HM, HEN, and HB) achieves the most balanced privacy-utility trade-off, with high robustness and moderate throughput and intelligibility (as indicated by circle diameter). The mid-ranked group (H2D, HP, and H3D) exhibits a moderate privacy-utility balance, with moderate robustness and intelligibility, but low throughput. In a lower-ranked group are HE and HS, with HE having moderate practicality performance across all factors.
In the lowest-ranked group are TSD, SPct, and LR+SR, primarily due to their poor utility. 


Overall, the experiments show that the proposed framework offers a comprehensive and interpretable approach to privacy-protection assessment by integrating multiple metrics. The HR-VISPR dataset improves the discriminability and interpretability of the privacy metric. Furthermore, the practicality is effectively assessed without extensive numerical analysis. While not a rigid benchmark for evaluating complex tasks under anonymization, the framework remains adaptable and generalizable.

\subsubsection{Ablation Study}
In the ablation tests, we analyzed the impact of weight selection on the practicality by varying the weights across robustness, intelligibility, and throughput. Emphasizing one or two weights highlights methods that achieve a balance in the corresponding factors. Results of the ablation study are shown in Table \ref{tab:ablation_weights}. In the first weight combination group, assigning a dominant weight to a single factor yields rankings similar to those shown in Fig. \ref{fig:3d-graph}, where the emphasized factor determines the overall practicality assessment. When robustness or throughput is emphasized, TSD, HM, and HP show the highest practicality. With intelligibility emphasized, HS achieves the highest score. In the second combination group, where two factors are equally weighted, methods that balance these factors best achieve the highest practicality. For instance, HM and HP rank highest when intelligibility and throughput are emphasized, while TSD and HM perform best when robustness and throughput are prioritized. Similarly, HM, HP, and HE achieve the highest practicality when robustness and intelligibility are equally emphasized. With equal weights, HM demonstrates the best balance across all factors.



\renewcommand{\arraystretch}{1}
\begin{table}[]
    \centering
    \caption{The impact of the weights on the practicality scores. Wr, Wi, Wt refer to the weights of robustness, intelligibility, and throughput, respectively.
    {\color{red}\textbf{Bold red}} = highest practicality;  
    }
    \label{tab:ablation_weights}
    \resizebox{\columnwidth}{!}{%
    \begin{tabular}{ccccccccccccc}
        \toprule
        \textbf{[Wr, Wi, Wt]} & \textbf{HB} & \textbf{HP} & \textbf{HE} & \textbf{HM} & \textbf{HEN} & \textbf{H2D} & \textbf{H3D} & \textbf{HS} & \textbf{LR+SR} & \textbf{SPct} & \textbf{TSD} \\
        \midrule
        $[0.8,0.1,0.1]$ & 
          0.87 & 0.88 & 0.87 & 0.88 & 0.87 & 0.73 & 0.29      & 0.10      & 0.86 & 0.84 & \best{0.91} \\
        $[0.1,0.8,0.1]$ & 
          0.71      & 0.75      & 0.73      &0.76      & 0.53      & 0.47      & 0.42      & \best{0.80} & 0.35      & 0.14      & 0.30      \\
        $[0.1,0.1,0.8]$ & 
          0.45      & 0.54      & 0.50      & 0.58      & 0.55      & 0.21      & 0.57 & 0.10      & 0.38      & 0.42      & \best{0.91} \\
        \midrule
        $[0.2,0.4,0.4]$ & 
          0.62      & 0.68      & 0.65      & \best{0.70}      & 0.59      & 0.40      & 0.46      & 0.40      & 0.44      & 0.36      & 0.65      \\
        $[0.4,0.2,0.4]$ & 
          0.67      & 0.72      & 0.69      & 0.73      & 0.68      & 0.47      & 0.43      & 0.20      & 0.58      & 0.56      & \best{0.83} \\
        $[0.4,0.4,0.2]$ & 
          0.74      & \best{0.78}      & 0.76      & \best{0.78}      & 0.68      & 0.55      & 0.39      & 0.40      & 0.57      & 0.48      & 0.65      \\
        \midrule
        $[0.3,0.3,0.3]$ & 
          0.67      & 0.72      & 0.69      & \best{0.73}      & 0.64      & 0.47      & 0.42      & 0.33      & 0.53      & 0.46      & 0.70      \\
        \bottomrule
    \end{tabular}
    }
\end{table}

\section{Discussion}\label{sec:discussions}

\noindent\textbf{HR-VISPR Limitations.} Even though the HR-VISPR dataset has improved the discriminability of the privacy attribute classifier, it remains limited by its broad labeling. For instance, it includes labels like weight, height, skin color, and casual clothing, which are inclusive and dominantly present. This labeling introduces a significant imbalance that complicates the model's learning, affecting its explainability. 


\noindent\textbf{Framework Limitations.} While the framework yields privacy scores aligned with human perception, it retains few limitations in privacy evaluation. Specifically, the privacy classifier does not produce optimal rankings and remains limited in explainability. For example, it ranks LR+SR as highly protective due to full-image corruption; however, human perception is powerful and can still recognize attributes, such as clothing, skin, and hair colors. Conversely, HS ranks the lowest, despite replacing humans with fake identities. Nevertheless, this can be mitigated by jointly analyzing privacy scores with the robustness metric, which reflects identity dissimilarity, see \textcolor{blue}{Supp. S4.}

\section{Conclusion}\label{sec:conclusions}
In this paper, we introduced a three-dimensional framework for evaluating human visual privacy-protection methods, focusing on privacy, utility, and practicality. At the core of this framework, we introduced a step towards an interpretable and discriminative privacy metric, trained on the proposed HR-VISPR human-centric dataset. HR-VISPR features comprehensive human-related labels, including biometric, soft-biometric, and non-biometric attributes. The framework serves as an insightful tool for comparing and contrasting anonymization techniques, while highlighting the sources of their limitations. We conducted experiments on a wide range of visual anonymization techniques, ranking them across the privacy, utility, and practicality dimensions. While achieving a perfect privacy-utility-practicality balance is unattainable, the framework provides valuable insights for guiding design decisions based on system requirements. Additionally, to facilitate trade-off analysis in diverse contexts, HR-VISPR, along with its anonymized versions and corresponding privacy and utility labels, is publicly available. This allows the framework to be extended to different contexts and more complex utility tasks. Additionally, HR-VISPR labels can be extended to video datasets to enable the evaluation of privacy protection in dynamic scenarios. 


{
    \small
    \bibliographystyle{ieeenat_fullname}
    \bibliography{references}
}

\end{document}